%% file: main.tex
\documentclass[runningheads]{llncs}

\usepackage{eccv}

\usepackage{eccvabbrv}

\usepackage{graphicx}
\usepackage{booktabs}
\usepackage{url}
\usepackage{svg}

\usepackage[accsupp]{axessibility}  %

\usepackage[hidelinks]{hyperref}

\usepackage{orcidlink}

\input{imports}

\input{definitions}

\begin{document}

\title{Stitched Embeddings: A Unified Latent Space for 3D Garments and 2D Patterns} 

\titlerunning{Stitched Embeddings}

\author{Andrea Sanchietti\inst{1,2} \and
Riccardo Marin\inst{3, 4} \and
Bharat Lal Bhatnagar\inst{6} \\
Yuanlu Xu\inst{6} \and
Gerard Pons-Moll\inst{1,2,5}
}

\authorrunning{A.~Sanchietti et al.}

\institute{\textsuperscript{1}University of T\"ubingen, Germany
\textsuperscript{2}T\"ubingen AI Center\\
\textsuperscript{3}Technical University of Munich, Germany
\textsuperscript{4}Munich Center for Machine Learning
\textsuperscript{5}Max Planck Institute for Informatics, Germany
\textsuperscript{6}Meta
}

\maketitle
{
\vspace{-0.9cm}

\begin{figure}
    \centering
    \scriptsize
    \captionsetup{type=figure}
    \newcommand{\teaserwidth}{\textwidth}
    
    \begin{overpic}[width=\textwidth, percent]{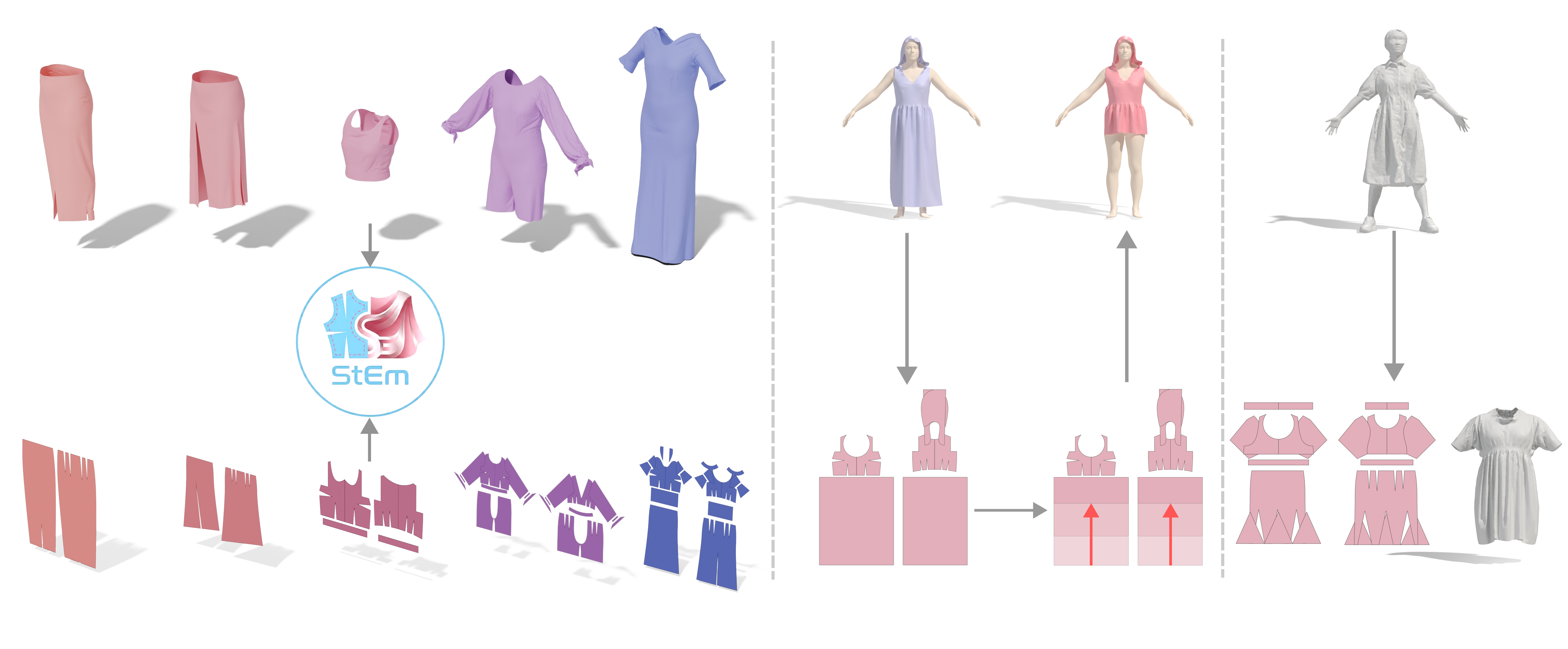}
        % Example of adding an overlay:
        \put(17,40){3D Garments}
        \put(16,3){Sewing Patterns}

        \put(54,40){Input}
        \put(53,3){Predict}
        \put(49,1){Sewing Patterns}
        \put(70,3){Edit}
        \put(68,40){Result}

        \put(87,40){Scan}
        \put(86,3){Predict}
        \put(82,1){Sewing Patterns}
    \end{overpic}

    \caption{\label{fig:teaser}
    In this work, we propose \textbf{Stitched Embeddings (StEm)}, a unified latent space to jointly represent 3D garments and their corresponding 2D sewing patterns. Our method, \method{}, provides state-of-the-art performance for sewing pattern prediction from 3D garments. Our approach is trained end-to-end, and supports test-time adaptation for sewing pattern predictions and seamless propagation of editing from sewing patterns to 3D realizations.}
\end{figure}
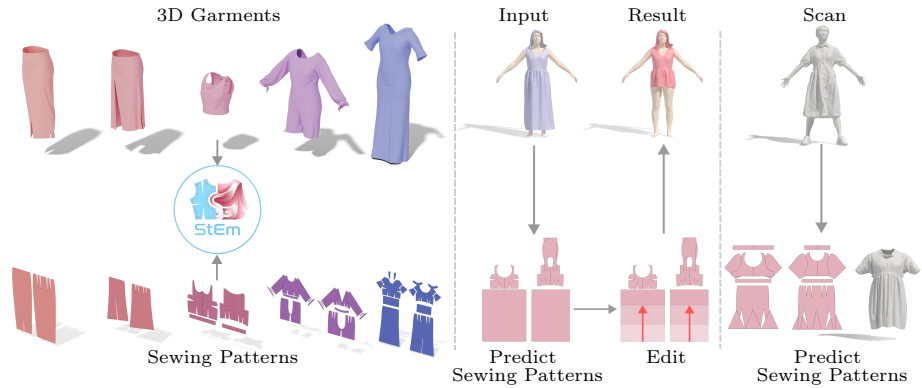

\vspace{-0.9cm}
\begin{abstract}
While garments are essential for realistic digital humans, their topological variety makes them much harder to model than parametric bodies. Traditional tailoring relies on 2D sewing patterns, yet bridging these patterns to 3D geometry currently requires physical simulations. We present Stitched Embeddings, the first simulation-free framework to unify 3D garment reconstruction and sewing pattern inference within a single bidirectional latent space. By leveraging the geometric priors of a pretrained 3D foundation model, our approach overcomes the data scarcity typically associated with high-quality garment modeling. We propose to use the BoxMesh as a critical intermediate representation to align 2D panels into 3D configurations without the computational overhead of a simulator. This architecture achieves state-of-the-art accuracy in pattern reconstruction while significantly improving efficiency. Furthermore, our differentiable pipeline enables novel applications, including pattern recovery from meshes and 3D editing from 2D patterns. Finally, this work provides a scalable link between neural 3D vision and the physical garment manufacturing pipeline. \href{https://andreus00.github.io/stitchedembeddings}{Project page: https://andreus00.github.io/stitchedembeddings}.

\end{abstract}

\section{Introduction}
\label{sec:intro}
\input{sections/01_Introduction.tex}

\section{Related Works}
\label{sec:related}
\input{sections/02_Related.tex}

\section{Background}
\label{sec:background}
\input{sections/03_Background.tex}

\section{Stitched Embedding Networks (\method{})}
\label{sec:method}
\input{sections/04_Method}

\vspace{-0.2cm}
\section{Results}
\label{sec:results}
\vspace{-0.35cm}
\input{sections/06_Results}

\vspace{-0.2cm}
\section{Conclusions}
\label{sec:conclusions}
\vspace{-0.2cm}
\input{sections/07_Conclusions.tex}

\bibliographystyle{splncs04}
\bibliography{main}

\clearpage
\newpage
% \appendix
% \input{sections/08_SupMat.tex}

\end{document}

%% file: imports.tex
\usepackage{epsfig}
\usepackage{graphicx}
\usepackage{comment}
\usepackage{amsmath,amssymb} %
\usepackage{hhline}
\usepackage[export]{adjustbox}
\usepackage{bbm}

\usepackage[font=small,skip=5pt]{caption}
\usepackage{float}

\usepackage{multirow}
\usepackage{tabularx}
\usepackage{soul}
\usepackage{array}
\usepackage{comment}
\usepackage{outlines}

\usepackage{overpic}
\usepackage{wrapfig}
\usepackage{pifont}
\usepackage{tikz}

%% file: definitions.tex
\newcommand{\method}[0]{StEm-Net}

\newcommand{\mypar}[1]{\noindent{\textit{\textbf{#1}}}}

%% file: sections/01_Introduction.tex
The rapid progress of 3D capture technology, together with advances in data-driven techniques, has opened unprecedented opportunities for human digitalization. Within this domain, garments constitute a fundamental pillar of individual identity and visual realism. But if Virtual Human research has provided dramatic advances thanks to the development of parametric body models, faithful garment reconstruction and manipulation have remained limited and far from being solved. The vast topological variability and non-rigid deformation of clothing make it mathematically intractable to reduce garments to a simple additive mesh model similar to SMPL \cite{SMPL:2015}. Hence, developments in neural fields have motivated scholars to represent them as unsigned distance fields (UDFs) \cite{he2023sketch2cloth, liu2024reconstructing, de2023drapenet,chibane2020ndf, vuran2025remu}.

However, implicit representations lack the semantic structure required for downstream applications in the tailoring and fashion industries. Manufacturing needs sewing patterns: a 2D, piece-wise CAD formalization that serves as the blueprint for production and realistic animations. Formally, these patterns consist of discrete 2D panels whose boundaries are defined by parametric curves (\eg, Bézier curves), interconnected via seam constraints. Traditionally, the bridge between 2D patterns and 3D geometry is established through physics-based simulation, in which panels are initialized in 3D space and "sewn" together using spring-mass systems. Researchers have collected several datasets \cite{NeuralTailor2022,GarmentCodeData:2024,antic2024close}, defined suitable tokenizations \cite{he2024dresscode, nakayama2025aipparel}, and achieved promising results in the prediction of sewing patterns from images \cite{can2026image2garment,bian2024chatgarment} or point clouds \cite{NeuralTailor2022}.

Despite all these efforts, we still observe a lack of scalability and generalization in the methods. Recent approaches aim to predict sewing patterns from 3D using a single-pass regression, followed by a physical simulation to obtain the final 3D geometry. However, physical simulators are slow and generally not differentiable, which prevents end-to-end learning and their use for optimization (\eg, recovery from incorrect predictions through test-time adaptation). More recently, differentiable physical simulators have become more affordable \cite{howell2022dojo,macklin2022warp,InverseGarment2024}. However, they remain slow and still do not support editing the number or type of sewing patterns during test-time optimization. Our research originates from a fundamental question: how can we combine the flexibility of the UDF representation with the controllability and application relevance of the sewing pattern? 

Our answer is \textbf{Stitched Embeddings}: a common latent space where 3D Garments and 2D Sewing Pattern Parameters coexist. We build such a space using a combination of 3D garments and BoxMesh, a representation in which sewing patterns are arranged in 3D as an initialization for a simulation. Our intuition is that it can serve as a spatial proxy that aligns 2D panels into a canonical 3D configuration according to their semantic function (\eg sleeves, front panel). By combining their spatial locations with their semantic meanings, we significantly facilitate learning a common latent space (see \cref{fig:teaser} for recovered samples from such a latent space). Our Stitched Embedding Network (\method{}) is built on top of a 3D foundational model, demonstrating for the first time the use of such backbones for 2D sewing patterns.

We measure our performance on two datasets of different complexity, showing that \method{} achieves state-of-the-art performance in sewing pattern prediction from 3D point clouds, improving metrics even up to 15\% over the closest competitor. Remarkably, although extracting semantics from 3D point clouds is more challenging than from 2D images, our method also surpasses 2D-based methods. In addition, the differentiability of our approach enables test-time latent optimization. We can further improve the network's initial prediction by backpropagating a reconstruction loss to yield sewing patterns that better align with the input 3D garment.
 Also, StEm allows us to edit one representation and observe the impact of the modifications on the other. For example, we can update the sewing pattern and obtain an updated 3D garment within the same framework, without requiring backpropagation or physical simulation.
Ultimately, this work represents a significant milestone towards a scalable, end-to-end learning framework that bridges the gap between neural 3D vision and the physical tailoring pipeline.

In summary, our contributions are:
\begin{enumerate}
    \item \emph{\method{}}: Our Stitched Embedding Network (\method{}) is the first end-to-end, fully differentiable, and simulation-free method that connects different garment representations in a unified latent space.
     \item We set a new state of the art in the task of sewing pattern prediction from 3D point clouds, reducing the error by up to $15\%$ compared to the closest competitor.
     \item We enable test-time adaptation and direct garment editing within the same framework, without the need for physical simulations, providing a practically flexible and useful tool for the Vision and Graphics communities.

\end{enumerate}
% Project page \href{https://andreus00.github.io/stitchedembeddings}{https://andreus00.github.io/stitchedembeddings}.

%% file: sections/02_Related.tex
\subsection{Garment Reconstruction}
Reconstructing 3D garments from disparate inputs (images, videos, noisy scans) has attracted broad interest \cite{pons2017clothcap, xiang2020monoclothcap, liu2024reconstructing, wang2018learning, eskandar2026dama, mir2026ahoy, kostyrko2026avaimg}, motivating significant efforts in dataset collections and annotation \cite{antic2024close, zhu2020deep, wang20244d, GarmentCodeData:2024, zou2023cloth4d, musoni2022gim3d, musoni2023gim3d, musoni2024capturing, garavaso2025point}, including procedural approaches \cite{KorostelevaGarmentData}. Despite the growing availability of such data, the space of possible garment shapes and deformations is vast and far from being fully covered. Hence, many methods reconstruct by relying on a fixed template for clothing \cite{zhu2020deep, zhang2017detailed} or a parametric template for the human body \cite{SMPL:2015, tiwari2021neural, tiwari2020sizer, Bhatnagar_2019_ICCV, antic2024close, li2023diffavatar}. On the other hand, such templates also constrain the garments that can be represented. Another characteristic of all the aforementioned methods is their use of Unsigned Distance Fields (UDF) as the main output for garments. Such a representation is convenient, as it allows flexible topology and continuous deformation. Other works combine 3D meshes and UDF fields to generate or reconstruct continuous surfaces, such as SurfD \cite{yu2023surf}, Ghost-on-the-Shell \cite{Liu2024gshell}, GarmentCrafter \cite{wang2025garmentcrafter}, and SimAvatar \cite{simavatar2024}. Recent works also find applications of Gaussian Splatting \cite{rong2025gaussian, liu2025clothedreamer, wang2025clocap, jung2025clothingtwin, Guo_2025_CVPR, gong2024laga}, while adapting them to practical use cases (\eg, modeling and animation) remains an open research area \cite{wu2024surface, wolf2024gs2mesh,zhang2025mega, wen2025intergsedit}. In this paper, we do not assume a fixed template or mesh, and instead propose connecting the output UDF to a parametric representation of garments, building on recent advances in sewing pattern modeling.

\subsection{Sewing Pattern Recovery}
\mypar{From Single Image.} In recent years, we have witnessed significant work on providing garments with an underlying parametric structure, also called sewing patterns \cite{pietroni2022computational, wolff2023designing, luo2025deep, shen2020gan, wolff2019reflection} (see \cref{fig:SP1}). Such a representation is borrowed from standard garment manufacturing, where clothes are obtained by designing small patches that are then sewn together. Sewing patterns are advantageous for applications and for bridging the gap between research and industry. Also, they provide a piecewise regularization, flexible enough to represent a vast set of garments, and also enforce physical plausibility. Since the sewing patterns naturally live on a 2D plane, a line of work tries to recover them and the 3D garments from such a flat representation \cite{yang2016detailed}. The work of Berthouzoz et al. \cite{berthouzoz2013parsing} was the first to propose a parsing from sewing pattern diagrams to 3D garment models, relying on a combination of machine learning and integer programming. More recently, Panelformer \cite{Chen_2024_WACV} and SewFormer \cite{liu2023sewformer} proposed reconstructing patterns from images by extracting features with a transformer and predicting sewing pattern panels and edges. Recent works such as GarmentX and DressCode \cite{guo2025garmentxautoregressiveparametricrepresentations, he2024dresscode} have explored autoregressive architectures to generate garment parameters from images. 
ChatGarment \cite{bian2024chatgarment} and AIpparel \cite{nakayama2025aipparel} finetune VLMs \cite{zhang2024vision} to generate patterns from text or image inputs. Design2GarmentCode \cite{zhou2025design2garmentcode} tackles the problem of predicting sewing patterns as a code-generation problem. GarmentImage \cite{tatsukawa2025garmentimage} uses a raster-based sewing pattern representation. SewingLDM \cite{liu2024multimodallatentdiffusionmodel} and GarmentDiffusion \cite{ijcai2025p163} explore methods to encode patterns in a compact latent space and then learn a diffusion process conditioned on input text or sketches. 
Though this line of work has been promising due to the limited input required, it is also highly underconstrained. With advances in acquisition and generation techniques, the availability of 3D models has become increasingly common \cite{xue2025gen, xue2024human, xiang2025structured, deitke2023objaverse, deitke2023objaverse2}. This motivated several methods to start directly from 3D assets.

\mypar{From 3D.} With the increased availability of 3D data, we witness an emergent attention toward recovering sewing patterns from 3D clothes. While early works were restricted to modeling simple garments such as skirts and t-shirts \cite{bang2021estimating,chen2015garment}, the more recent NeuralTailor \cite{NeuralTailor2022} reconstructs sewing patterns from 3D point clouds and trains on the GarmentData \cite{KorostelevaGarmentData} dataset, which has a lower number of patterns than the GarmentCodeData \cite{GarmentCode2023, GarmentCodeData:2024}, and predicts patterns by regressing edges and stitches of panels with LSTMs~\cite{hochreiter1997long}. The main downside of NeuralTailor is its reliance on a final garment mesh simulation, which hinders the possibility of backpropagating through the network to optimize the pattern predictions. GarmageNet \cite{li2025garmagenet} introduces a unified 2D–3D garment representation that encodes each sewing-pattern panel as a geometry image \cite{gu2002geometry} in UV space, jointly capturing panel contours and draped 3D geometry. These panel-wise representations are compressed into a latent space and generated jointly using a diffusion transformer. Their method enables pattern-based reconstruction of garments from images and point clouds. In a concurrent work, named Garment Particles \cite{nakayama2026garment}, authors propose to diffuse a 5D point cloud that jointly models 3D geometry and 2D patterns. Another concurrent work Inverse Draping \cite{11474955} proposes formulating the pattern prediction from meshes as an autoregressive BoxMesh generation. Finally, ReWeaver \cite{li2026reweaver} is a contemporary work that starts from multi-views, uses VGGT \cite{wang2025vggt} to recover a point cloud, and then predicts the SP parameters. None of the previous methods provides an end-to-end approach to go from 3D to a sewing pattern and vice versa, and they require complicated machinery to combine 3D and sewing patterns. This makes it impossible to run any optimization on top of them.

%% file: sections/03_Background.tex
\begin{figure}[t!]
    \centering
    \scriptsize
    % Left Image
    \begin{minipage}{0.54\textwidth}
        \begin{overpic}[width=\linewidth]{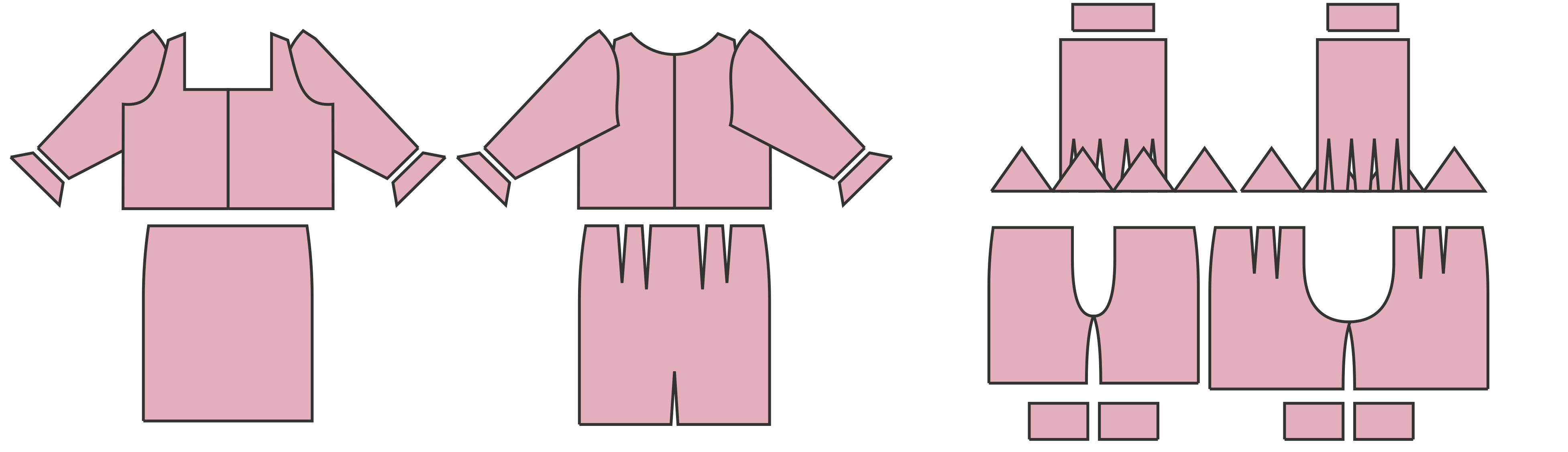}
        
            \put(10,29.5){Sewing Patterns \#1}
            \put(60,29.5){Sewing Patterns \#2}
        \end{overpic}
        \caption{An example of sewing patterns. Every garment is represented by a set of panels. While sometimes their semantic meaning is clear (\eg, for the body), for others it is difficult to guess without simulating them (\eg, the pieces on the right are a skirt and pants).} 
        \label{fig:SP1}
    \end{minipage}
    \hfill
    % Right Image
    \begin{minipage}{0.38\textwidth}
    \vspace{-0.6cm}
        \begin{overpic}[width=\linewidth]{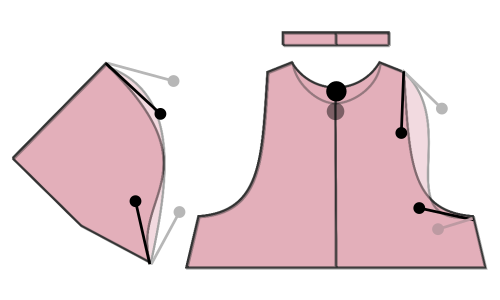}
        \end{overpic}
        \vspace{-0.6cm}
        \caption{An example of Sewing Pattern variability. Every piece is defined by B\'{e}zier curves; their shape can be modified by acting on the control points.}
        \label{fig:SP2}
    \end{minipage}
\end{figure}

\subsection{Sewing Pattern Representation}
\label{sec:SewingPatterns}

In traditional garment modeling, sewing patterns are defined as a collection of 2D boundary curves (typically a mixture of linear segments and quadratic or cubic B\'{e}zier curves, see \cref{fig:SP1,fig:SP2}) representing flat fabric panels. Formally, a pattern can be described by a set of vertices $V \subset \mathbb{R}^2$, edges $E$, and a set of curvature parameters $C$ that define the spline geometry. While this representation is precise for manufacturing, it is high-dimensional, non-uniform in length, and difficult for standard regression networks to predict directly.

To overcome this, we leverage the parametric space of GarmentCode \cite{GarmentCode2023}. Instead of predicting individual B\'{e}zier control points, our model operates on a fixed-dimensional set of \textbf{design parameters} (\textbf{122} for GarmentCodeData \cite{GarmentCodeData:2024}). These parameters provide a high-level, low-dimensional abstraction of the garment (\eg sleeve length, waist width, or neckline style) from which the full geometric sewing patterns can be deterministically generated.

To bridge the gap between 2D patterns and 3D geometry, we utilize a proxy representation named the \textbf{BoxMesh}. The BoxMesh is generated using GarmentCode, which places the 2D patterns in 3D space around a body model and applies a preliminary meshing procedure. It maintains the exact topology of the sewing patterns but occupies a 3D volume. When a cloth simulation is applied to the BoxMesh, it yields the final deformed \textbf{Garment Mesh}. By using the BoxMesh as an input modality, we provide our network with 3D spatial priors that are topologically aligned with the final output.

\subsection{Hunyuan3D's VAE Architecture}
\label{sec:Hunyuan}

Our architecture is built on top of the Hunyuan3D 2.1 \cite{hunyuan3d2025hunyuan3d} VAE framework, which we modify to better suit our task. While the original work utilizes Signed Distance Fields (SDF) to model watertight objects, we adopt an \textbf{Unsigned Distance Field (UDF)}. This allows our model to represent garments as non-watertight, open manifolds, which is essential for capturing realistic clothing structures like necklines and open hems.

A critical challenge in our cross-modal task is ensuring that the latent space is invariant to whether the input is a BoxMesh or a simulated Garment Mesh. Since the encoder processes meshes by sampling point clouds via Farthest Point Sampling (FPS), the resulting latent sets are inherently unordered. To ensure feature alignment, we exploit the fact that the BoxMesh and the Garment Mesh share an identical topology. During training, we sample \textbf{corresponding points} from both geometries. This ensures that the same physical locations on a garment, whether in its pre-simulated ``box'' state or its final draped state, map to the same regions in the latent space.

%% file: sections/04_Method.tex
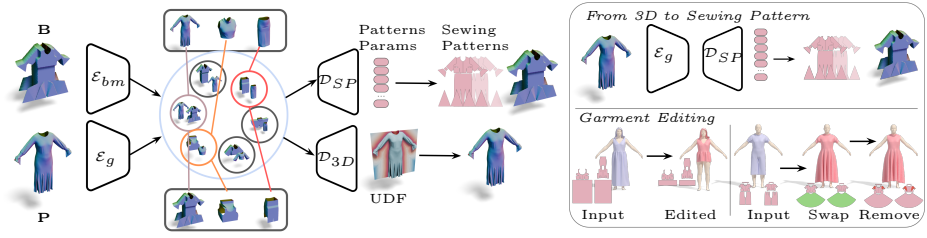
\begin{figure*}[!t]
\tiny
\centering
\begin{minipage}{0.6\linewidth}
    \centering
    \begin{overpic}[trim=0cm 0cm 0cm 0cm,clip,width=\linewidth]{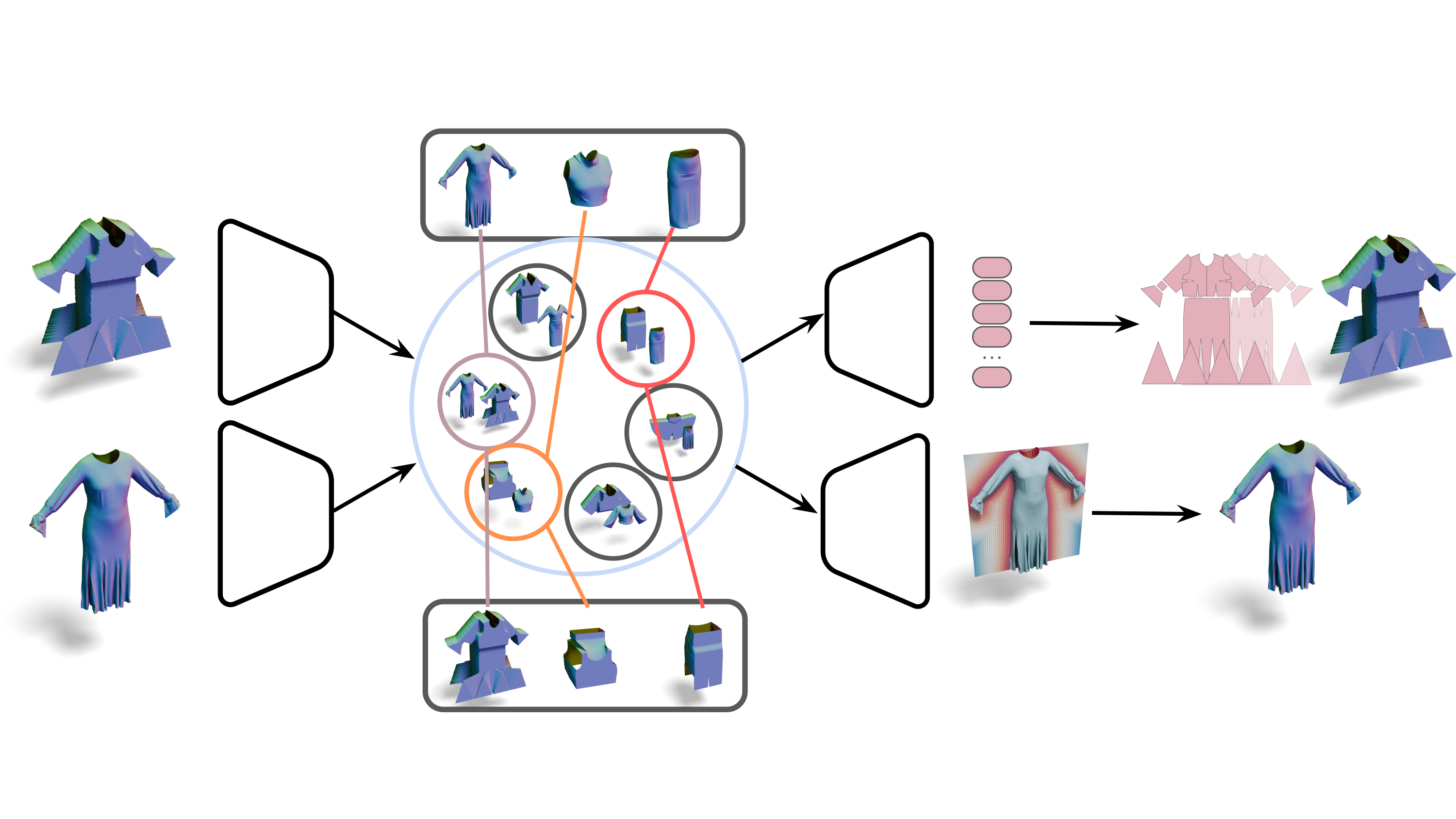}
    \put(6, 43){$\mathbf{B}$}
    \put(6, 9){$\mathbf{P}$}

    \put(15.7, 34.5){$\mathcal{E}_{bm}$}
    \put(16.5, 20){$\mathcal{E}_{g}$}

    \put(56.6, 34){$\mathcal{D}_{SP}$}
    \put(56.5, 20){$\mathcal{D}_{3D}$}

    \put(64, 43){Patterns}
    \put(64, 40){Params}

    \put(79, 43){Sewing}
    \put(79, 40){Patterns}

    \put(66, 12.5){UDF}

    \end{overpic}
\end{minipage}
\hfill
\begin{minipage}{0.39\linewidth}
    \centering
        \begin{overpic}[trim=0cm 0cm 0cm 0cm,clip,width=\linewidth, tics=5]{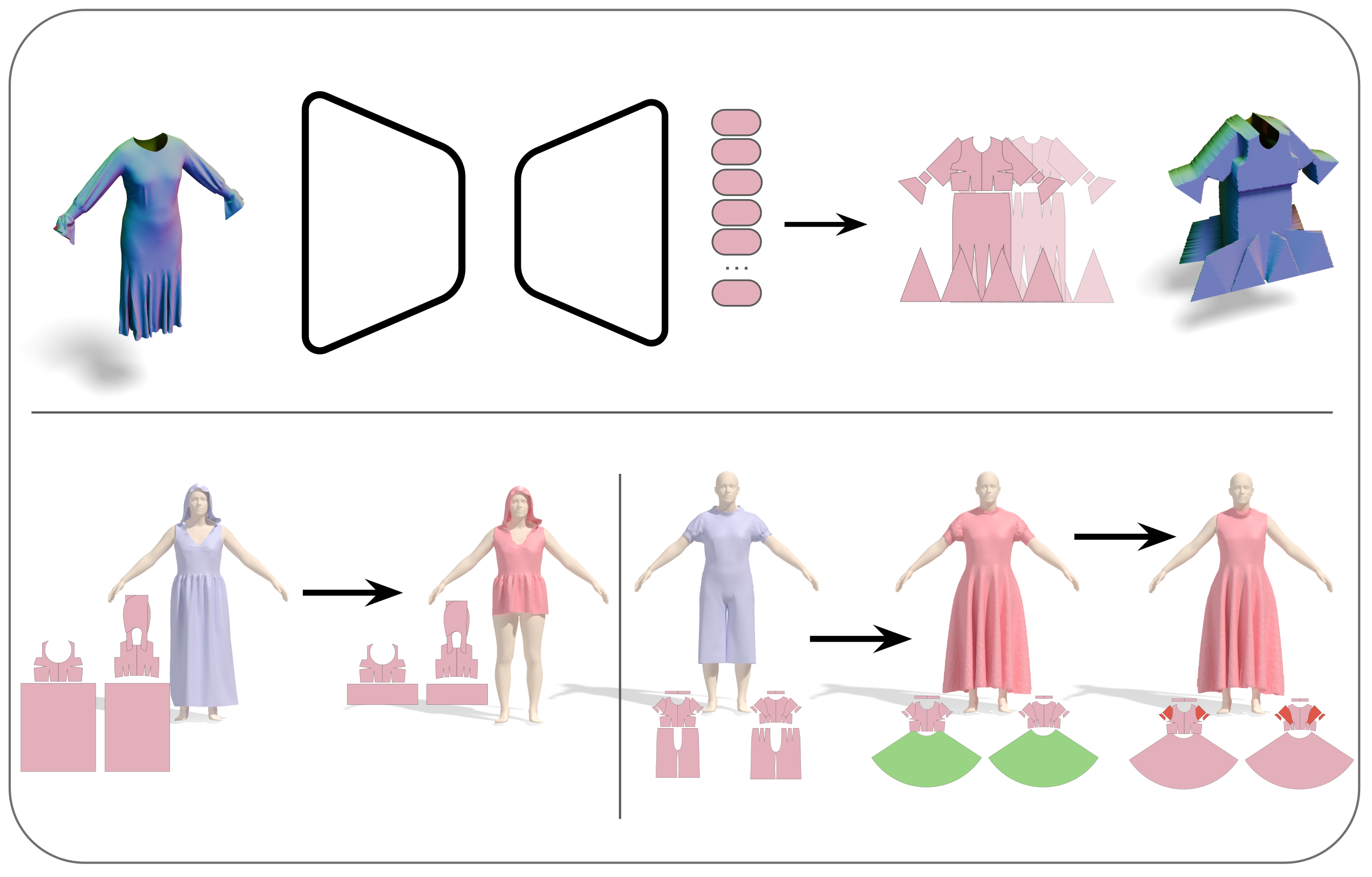}
    \put(5, 58.5){{\textit{From 3D to Sewing Pattern}}}

    \put(24, 48){$\mathcal{E}_{g}$}
    \put(37.8, 48){$\mathcal{D}_{SP}$}

    \put(3, 30){{\textit{Garment Editing}}}
    \put(4, 3){Input}
    \put(27, 3){Edited}

    \put(50, 3){Input}
    \put(67, 3){Swap}
    \put(81, 3){Remove}
    
    \end{overpic}
    
\end{minipage}

\caption{\label{fig:method}
On the left: \method{} is the first end-to-end differentiable method that unifies multiple different representations in a common compact representation. The model can map BoxMeshes and 3D garments into Stitched Embeddings. From such a latent space, we decode sewing pattern parameters and UDFs. On the right, we show our method's modalities: prediction of Pattern Parameters from an input Mesh (top), and editing of 3D Garments based on 2D edits (bottom).}
\end{figure*}

\mypar{Overview.} In this section, we describe \method{}, the first learning-based approach differentiable end-to-end, that unifies 3D garment realizations and sewing patterns in a common shared latent space. We call such a representation Stitched Embeddings (StEm), and in the following, we describe how we design its encoding and decoding. Next, we outline the training procedure of \method{}. We present our pipeline in Fig. \ref{fig:method}.

\mypar{StEm encoding.} \label{par:encoding} Our goal is to define a common, grounding representation, where UDFs and sewing patterns are paired. Such a task is far from trivial: UDFs are 3D continuous representations, well modeled by parametric functions such as neural fields. Sewing Patterns are a discrete set of 2D pieces, parametrized by a continuous boundary. To date, we are unaware of any method capable of providing a shared latent space for these two different representations. To solve this, we propose Stitched Embeddings: a latent space that combines 3D garments with the BoxMesh of their sewing patterns, providing spatial connections to the 3D UDFs while also expressing the semantics of the 2D sewing patterns. To do so, we rely on two encoders: one for the 3D garments ($\mathcal{E}_{g}$) and one for the pre-simulation BoxMesh ($\mathcal{E}_{bm}$). The garment encoder $\mathcal{E}_{g}$ takes as input a garment represented as a point cloud $\mathbf{P} \in \mathbb{R}^{5120 \times 3}$ equipped with normals $\mathbf{N} \in \mathbb{R}^{5120 \times 3}$. The BoxMesh encoder $\mathcal{E}_{bm}$ processes the pre-simulated geometry. At training time, in order to align the latent distribution of the BoxMesh encoder with that of the garment encoder, we exploit the known $1:1$ correspondence between the simulated garment and its pre-simulation BoxMesh. Instead of independently sampling points from the BoxMesh surface, we construct a point cloud $\mathbf{B} \in \mathbb{R}^{5120 \times 3}$ such that each point in $\mathbf{B}$ corresponds exactly to a point in $\mathbf{P}$. To ensure consistency of the input representation, we associate with $\mathbf{B}$ the same normals $\mathbf{N}$ used for the garments. 
In our implementation, both encoders are inherited from Hunyuan3D \cite{hunyuan3d2025hunyuan3d} and map to a common latent set $\mathbf{z} \in \mathbb{R}^{256 \times 64}$. 

\mypar{StEm decoding.} We are interested in two representations: the UDF of the 3D garments positioned in space, and the related sewing patterns. These two are the most useful in practice: the UDF provides a 3D continuous representation that can be optimized to fit observations such as scans, while sewing patterns are fundamental for manufacturing (\eg, garment editing) and graphics pipelines (\eg, animation). Hence, the 3D decoder $\mathcal{D}_{3D}$ reconstructs the UDF field (truncated at $0.01$), while the sewing pattern decoder $\mathcal{D}_{SP}$ regresses the $122$ design parameters. We decode these parameters from the latent set with a decoding transformer, after which learnable per-panel queries cross-attend over the full latent sequence, yielding one feature per panel that lightweight MLP heads map to that panel's parameters. These parameters correspond exactly to the parametric design space defined by the underlying garment modeling system \cite{GarmentCode2023}, which deterministically generates sewing patterns from these structured parameters. Consequently, our prediction space aligns with the expressive range of the parametric generator, ensuring valid, manufacturable sewing patterns by construction.
To ensure the model remains body-aware, all meshes are normalized relative to the transformation required to fit the underlying body model into a $[-1, 1]$ unit cube.

\subsection{\method~Training Losses.}
Our architecture requires the model both to produce reliable reconstructions and to learn a latent representation in which sewing patterns and 3D models of the same garment are close together. Hence, we optimize \method~with a composite loss function:
\begin{equation}
    \mathcal{L} = \mathcal{L}_{UDF} + \lambda_{1}\mathcal{L}_{SP} + \lambda_{2}\mathcal{L}_{KL}
\end{equation}
Where $\mathcal{L}_{UDF}$ is a 3D reconstruction loss, $\mathcal{L}_{SP}$ is a sewing pattern loss, and $\mathcal{L}_{KL}$ is used to enforce latent alignment.

\mypar{3D Reconstruction Loss} 
The UDF loss minimizes the Mean Squared Error (MSE) between the predicted and ground truth distance fields for both garment and BoxMesh inputs:
\begin{align}
    \mathcal{L}_{\text{UDF}} & = \sum_{p\in \Omega} \text{MSE}(\mathcal{D}_{3D}(\mathcal{E}_{g}(\mathbf{P}, \mathbf{N}), p), \text{UDF}(p)) \\ & +   \sum_{p\in \Omega} \text{MSE}(\mathcal{D}_{3D}(\mathcal{E}_{bm}(\mathbf{B}, \mathbf{N}), p), \text{UDF}(p))
\end{align}
where $p$ represents points sampled within the volume $\Omega = \left[-1, 1\right]^3 \subset \mathbb{R}^3$, and $\text{UDF}(p)$ returns the ground truth distance of $p$ from the surface. We adopt a stratified sampling strategy, by sampling half of the points randomly, and half at varying distances (\ie, $\pm0.0015$, $\pm0.005$, $\pm0.025$, $\pm0.05$) from the surface to capture both high-frequency details and the global field structure.

\mypar{Sewing Pattern Loss} 
The $122$ design parameters are heterogeneous, comprising continuous, boolean, and categorical values. We therefore assign each parameter to a prediction head, and split $\mathcal{L}_{SP}$ into:
\begin{equation}
    \mathcal{L}_{SP} = \sum_{i \in \text{cont}} \text{MSE}(y_i, \hat{y}_i) + \sum_{j \in \text{bool}} \text{BCE}(y_j, \hat{y}_j) + \sum_{k \in \text{cat}} \text{CE}(y_k, \hat{y}_k)
\end{equation}
where $\text{MSE}$ handles continuous dimensions (\eg panel dimensions), $\text{BCE}$ handles binary flags (\eg sleeve presence), $\text{CE}$ handles multi-class categories (\eg, garment type), $y_n$ represents the output of the $n$-th prediction head, and $\hat{y}_n$ represents the ground truth value of the $n$-th parameter.

\mypar{Latent Alignment} 
The KL-divergence loss serves a dual purpose: regularizing the latent space against a standard normal distribution and enforcing consistency between the two input modalities:
\begin{equation} \label{eqn:kl_losses}
    \mathcal{L}_{KL} = \text{KL}(z_{G} \parallel \mathcal{N}(0, I)) + \text{KL}(z_{B} \parallel z_{G})
\end{equation}
This formulation forces the BoxMesh encoding $z_B$ to reside in the same manifold as the garment encoding $z_G$, allowing the model to infer 2D patterns from 3D scans effectively.

\subsection{\method~Inference}
\method{} enables bidirectional mapping between the 3D geometric domain and the 2D sewing pattern space.

\mypar{From 3D to Sewing Patterns} 
For inference from 3D observations, we encode a point cloud $\mathbf{P}$ and normals $\mathbf{N}$ through $\mathcal{E}_{g}$ to obtain its representation in the StEm latent space. From this shared embedding, the decoder $\mathcal{D}_{SP}$ regresses structured design parameters, reconstructing the 2D manufacturing specifications directly from the 3D geometry.
While our encoders provide a robust initialization, the mapping from unseen 3D geometries to the latent space is not always lossless. To bridge this domain gap and refine the predicted sewing patterns, we perform \emph{test-time latent optimization}. Given an input mesh, we initialize its latent vector set $\mathbf{z}$ and iteratively refine it by minimizing the reconstruction error between the predicted UDF and the input surface:
\begin{equation}
    \mathbf{z}^{*} = arg\ \underset{\mathbf{z}}{min} \sum_{p\in \Omega} \text{MSE}(\mathcal{D}_{3D}(\mathbf{z}, p), \text{UDF}(p))
\end{equation}
where $\Omega \subset \mathbb{R}^3$ represents the set of points sampled within the bounding volume, $\mathcal{D}_{3D}(\mathbf{z}, p)$ is the distance predicted by the 3D decoder at point $p$ given the latent set $\mathbf{z}$ and $\text{UDF}(p)$ is the ground truth distance from the input surface.

\begin{figure}[t!]
    \centering
    \begin{overpic}[trim=0cm 0cm 0cm 0cm,clip,width=\linewidth]{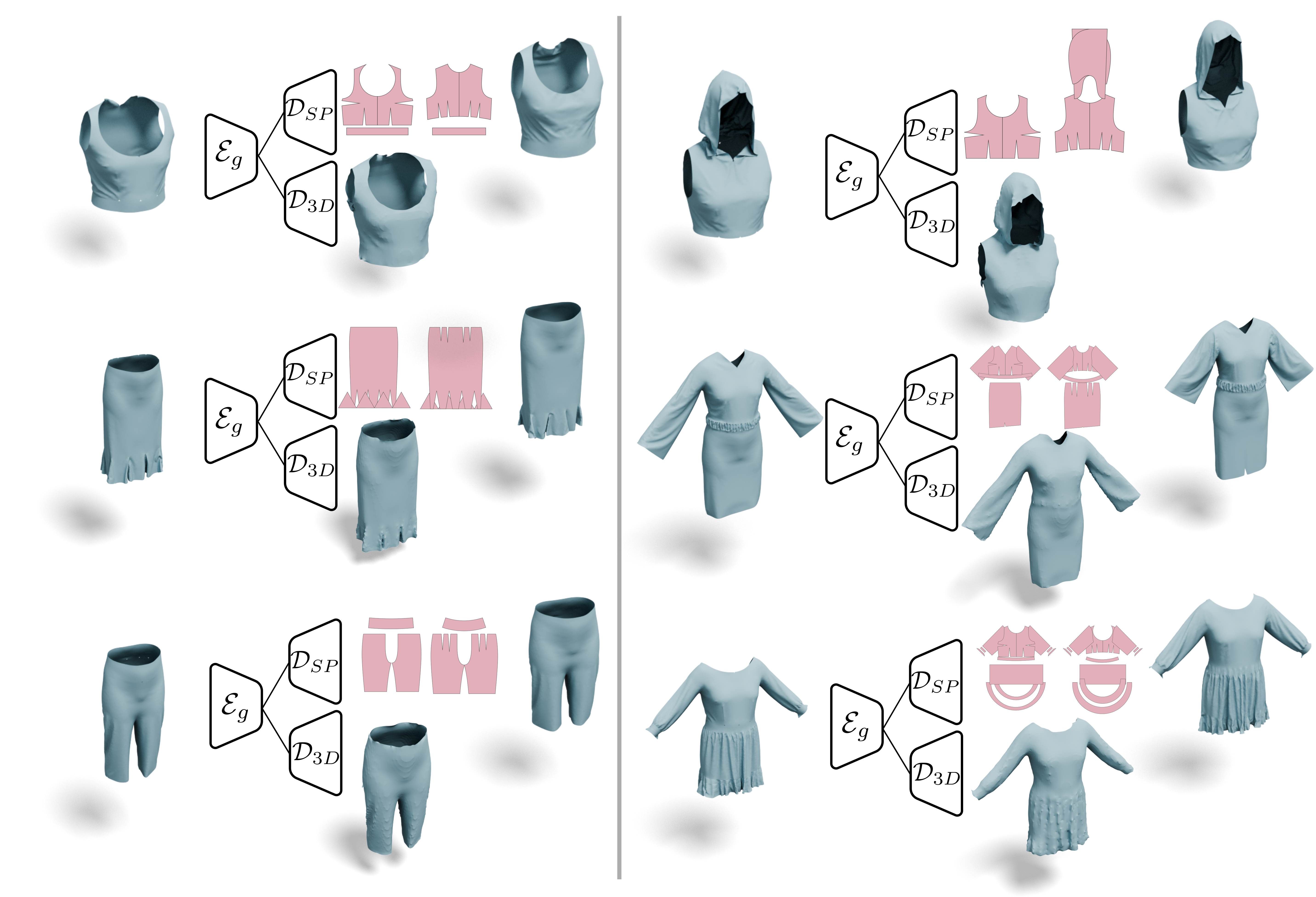}    
    \end{overpic}
    
    \caption{Autoencoding of \method. On the left, we show the input mesh. We pass this mesh to our garment encoder $\mathcal{E}_g$, which maps it into our StEm latent space. From that, we recover the 3D garment via $\mathcal{D}_{3D}$, and the sewing patterns via $\mathcal{D}_{SP}$, which we also simulate to obtain the corresponding 3D. We can appreciate that both the 3D output and the simulated sewing patterns produce geometry close to the ground truth, proving the reliability of our approach.}
    \label{fig:autoencoding}
\end{figure}

StEm is the first approach whose test-time optimization can modify both the continuous geometry and the discrete structure of a sewing pattern. Because the StEm latent space is shared and differentiable, the gradient-based refinement of the 3D geometry directly propagates to the sewing pattern parameters. This allows the model to ``correct'' the 2D design in response to the 3D observations, ensuring the final output is not just a static prediction but an optimized fit. Our experiments demonstrate that this refinement improves the accuracy of the regressed design parameters, particularly for out-of-distribution garments.

In Fig. \ref{fig:autoencoding}, we show the result of autoencoding the input mesh while predicting the pattern. To validate these results, we computed the mean Chamfer Distance between the output of $\mathcal{D}_{3D}$ and the simulation result obtained from $\mathcal{D}_{SP}$’s output, yielding a value of 5.5 cm.

\mypar{From Sewing Patterns to 3D} Because the mapping from a 2D pattern to 3D is inherently one-to-many (as garments can drape in various states), models trained only on geometry often produce over-smoothed results. To resolve this ambiguity, we condition our encoding on surface normals $\mathbf{N}$. These directional priors transform the ill-posed $1:N$ mapping into a deterministic $1:1$ correspondence, allowing the network to capture high-frequency folds and generalize to unseen drapes. \method{} also functions as a high-speed ``neural simulator''. By processing a pre-simulation BoxMesh through $\mathcal{E}_{bm}$, the model predicts the corresponding 3D UDF field in a fraction of a second, bypassing traditional physics-based solvers.

%% file: sections/06_Results.tex
\mypar{Baselines.} We evaluate our method against NeuralTailor \cite{NeuralTailor2022}, as it represents the most closely related approach to end-to-end sewing pattern estimation. NeuralTailor utilizes a PointNet-based encoder to process input scans and regresses pattern parameters through an LSTM-based decoder. To isolate the impact of our refinement process, we also include an ablation of our model without test-time optimization. While we attempted to train NeuralTailor on the more complex GarmentCodeData, the model failed to converge. We attribute this to the significant increase in structural complexity: GarmentCodeData\cite{GarmentCodeData:2024} contains up to 75 panels per garment (compared to 24 in the original GarmentData) and features a variable number of edges per panel, whereas the NeuralTailor dataset assumes a fixed topology. Furthermore, the significantly larger scale of GarmentCodeData likely exceeded the representational capacity of the baseline’s sequential decoding architecture. Consequently, we report comparative results on the original GarmentData \cite{NeuralTailor2022} to ensure a fair evaluation. For training and testing, we used the public train-test splits for the datasets.

\begin{table}[t]
\centering
\scriptsize
\setlength{\tabcolsep}{3pt}

\begin{adjustbox}{width=\linewidth}
\begin{tabular}{lccccc|ccccc}
\toprule
& \multicolumn{5}{c}{NeuralTailor Testset} 
& \multicolumn{5}{c}{GarmentCodeData TestSet} 
 \\

Method 
& Panel L2 ↓ 
& \#Panels 
& \#Edges 
& Stitch Precision
& Stitch Recall
& Panel L2 ↓ 
& \#Panels 
& \#Edges 
& Stitch Precision
& Stitch Recall\\
\midrule

NeuralTailor\cite{NeuralTailor2022} & 5.2 & 83.6\% & 87.3\% & 74.7\% & 83.9\%  & - & - & - & - & - \\
\method{} & 4.6 & 81.6\% & 93.3\% & 85.7\% & 86.0\%  & 4.4 & 69.5\% & 82.6\% & 66.6\% & 66.4\% \\
\method{} + Optim & \textbf{4.4} & \textbf{84.6\%} & \textbf{95.0\%} & \textbf{88.5\%} & \textbf{89.2\%}  & 4.3 & 71.1\% & 80.6\% & 64.0\% & 64.3\% \\

\bottomrule
\end{tabular}
\end{adjustbox}

\caption{Comparison of NeuralTailor and GarmentCodeData test sets. On NeuralTailor, our method provides better performance both before and after the optimization step. On GarmentCodeData, \method~provides a similar Panel L2 error, while NeuralTailor fails to converge.}
\label{tab:results}
\end{table}

\subsection{Comparison with 3D-based Baselines}
\begin{figure*}[!t]
\tiny
 \begin{overpic}[trim=0cm 0cm 0cm 0cm,clip, width=\linewidth]{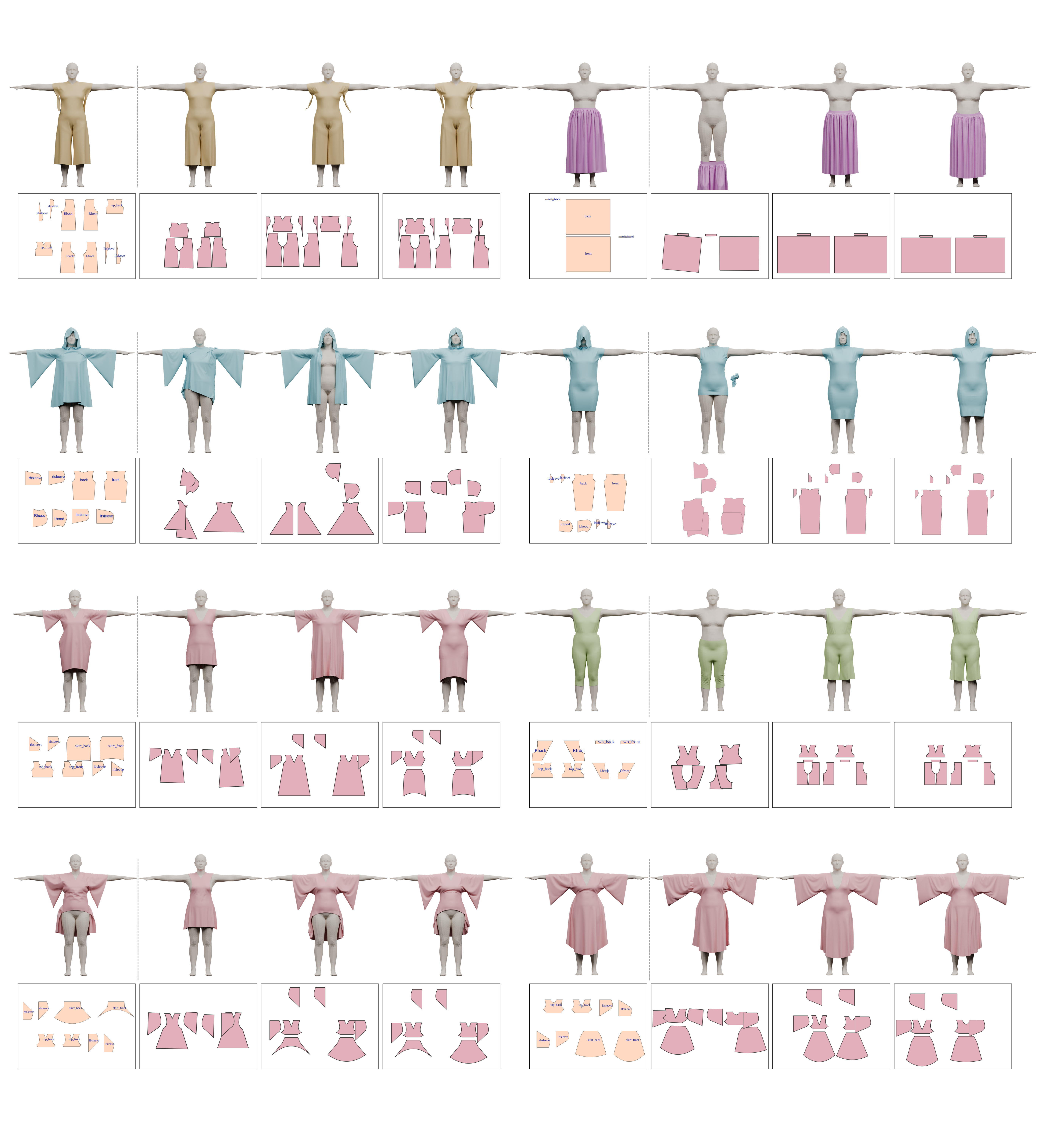}
 
    \put(45,1){\color{black!40}\line(0,1){90}}

    \put(4,95){Input}
    \put(5,74.5){GT}
    \put(12,74.5){NeuralTailor}
    \put(15,73){~\cite{NeuralTailor2022}}
    \put(24,74.5){\textbf{\method}}
    \put(34,74.5){\textbf{\method}}
    \put(34,73){+Optim}

    \put(4,72){Input}
    \put(5,51.5){GT}
    \put(13,51.5){NeuralTailor}
    \put(16,50){~\cite{NeuralTailor2022}}
    \put(24,51.5){\textbf{\method}}
    \put(34,51.5){\textbf{\method}}
    \put(34,50){+Optim}

    \put(4,49){Input}
    \put(5,28.5){GT}
    \put(13,28.5){NeuralTailor}
    \put(16,27){~\cite{NeuralTailor2022}}
    \put(24,28.5){\textbf{\method}}
    \put(34,28.5){\textbf{\method}}
    \put(34,27){+Optim}

    \put(4,26){Input}
    \put(5,5.5){GT}
    \put(13,5.5){NeuralTailor}
    \put(16,4){~\cite{NeuralTailor2022}}
    \put(24,5.5){\textbf{\method}}
    \put(34,5.5){\textbf{\method}}
    \put(34,4){+Optim}

 %-----------------

    \put(49,95){Input}
    \put(50,74.5){GT}
    \put(57,74.5){NeuralTailor}
    \put(60,73){~\cite{NeuralTailor2022}}
    \put(69,74.5){\textbf{\method}}
    \put(80,74.5){\textbf{\method}}
    \put(80,73){+Optim}

    \put(49,72){Input}
    \put(50,51.5){GT}
    \put(58,51.5){NeuralTailor}
    \put(61,50){~\cite{NeuralTailor2022}}
    \put(69,51.5){\textbf{\method}}
    \put(79,51.5){\textbf{\method}}
    \put(79,50){+Optim}

    \put(49,49){Input}
    \put(50,28.5){GT}
    \put(58,28.5){NeuralTailor}
    \put(61,27){~\cite{NeuralTailor2022}}
    \put(69,28.5){\textbf{\method}}
    \put(79,28.5){\textbf{\method}}
    \put(79,27){+Optim}

    \put(49,26){Input}
    \put(50,5.5){GT}
    \put(57,5.5){NeuralTailor}
    \put(61,4){~\cite{NeuralTailor2022}}
    \put(69,5.5){\textbf{\method}}
    \put(79,5.5){\textbf{\method}}
    \put(79,4){+Optim}

 \end{overpic}
\caption{\label{fig:qualitative} Qualitative comparison on NeuralTailor test set. Our method provides robust predictions, even in the presence of nuances such as strings or hoodies with long, large sleeves. NeuralTailor\cite{NeuralTailor2022} predicts inaccurate patterns, often resulting in dramatic failures. Our test-time latent optimization often recovers nuances and fixes inconsistencies, whereas in the worst case, it preserves the quality of the original prediction.}

\end{figure*}
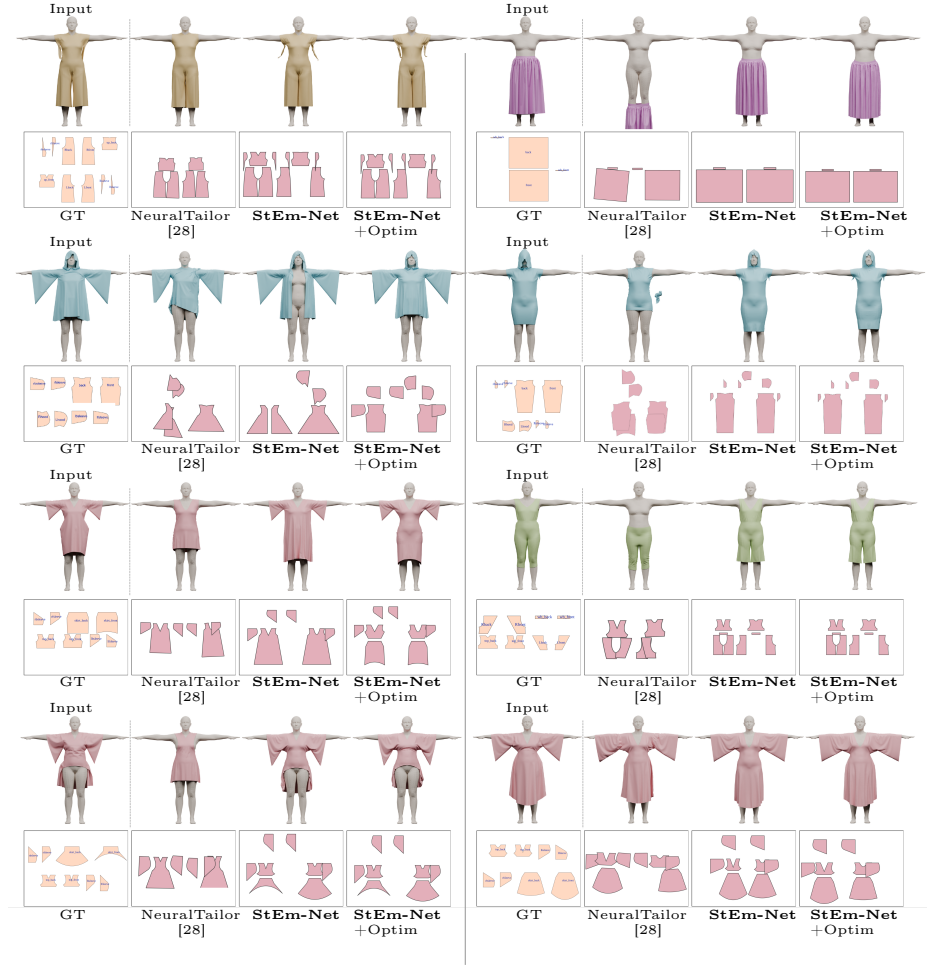

We evaluate our method on the task of estimating sewing patterns from 3D point clouds. We compare \method~against NeuralTailor\cite{NeuralTailor2022}, which, to date, is the only method with available code that solves an analogous task. As data, we consider the NeuralTailor and GarmentCode test sets, and in \cref{tab:results} we report our results. For the NeuralTailor, we used their pretrained checkpoint for comparison and trained our method on their training set. We observe that our method outperforms the baseline across almost all metrics, even before the optimization procedure. Particularly significant is the gain in the number of edges and stitch precision, demonstrating that our method is much more accurate in understanding the correct shapes of the sewing patterns and their interconnection. Likely, the use of BoxMesh provides better semantic supervision of individual patches during learning. After test-time optimization, we achieve even further improvements, with around 15\% better L2 reconstruction and almost 15\% better stitching precision. We also tried to compare ourselves on GarmentCode, but our attempts to train NeuralTailor on such a dataset failed to converge. Such a failure is likely caused by GarmentCode's complexity. Our model, on the other hand, converges successfully. Hence, we provide the first method that actually works on GarmentCode, offering an operational baseline for this dataset. To provide a better intuition for the quality of our predictions, we present several qualitative results in \cref{fig:qualitative}. For each sample, we report the input 3D garment and the ground-truth sewing patterns on the left, while on the right, we show the sewing pattern predictions of the methods and the 3D representation generated by physical simulation. Our method achieves significantly more accurate results.
\vspace{-0.05cm}
\subsection{Comparison with Image-based Baselines}

\begin{figure*}[!t]
\centering
    \begin{overpic}[trim=0cm 0.5cm 0.0cm 0.0cm,clip,width=\linewidth]{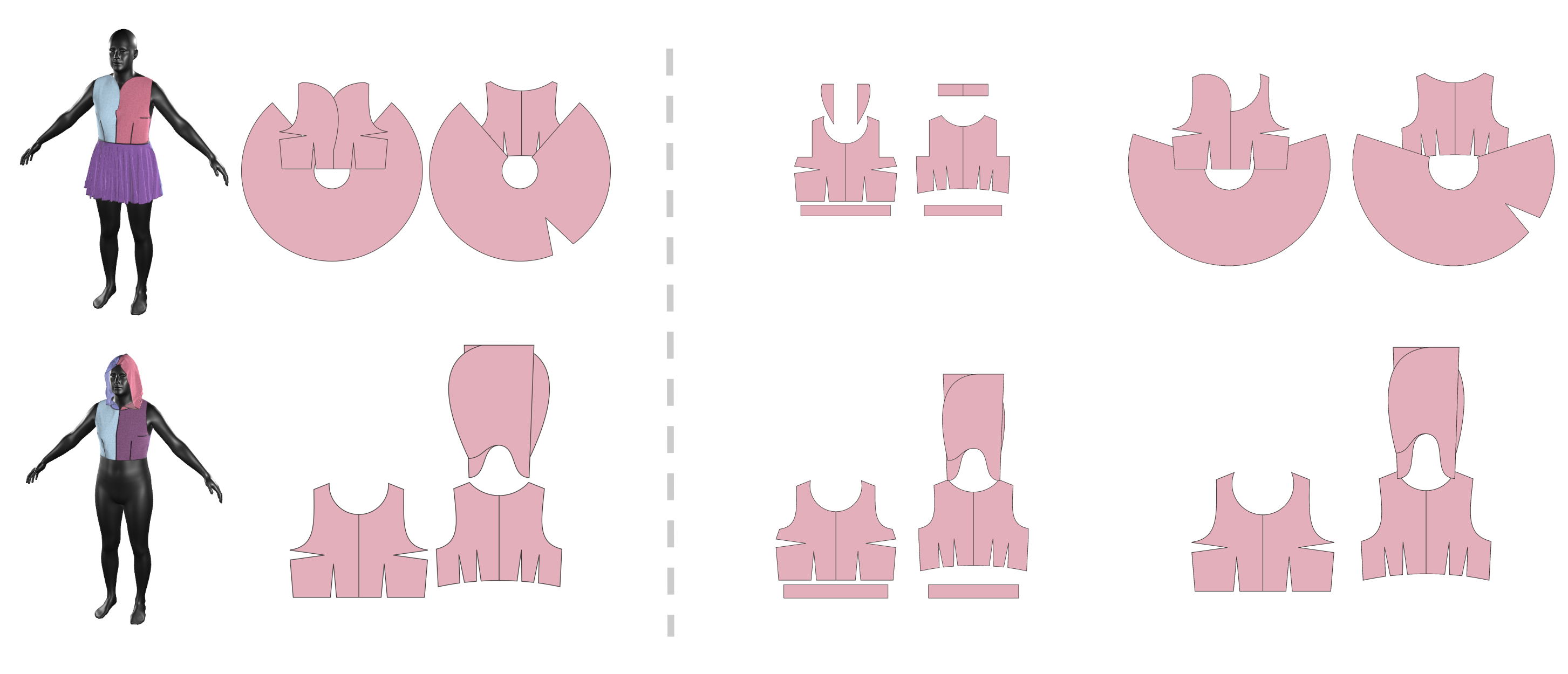}
\put(4, 40){Input}
\put(25, 40){GT}

\put(49, 40){ChatGarment~\cite{bian2024chatgarment}}
\put(82, 40){\textbf{\method{}}}

       \end{overpic}
\caption{\label{fig:qualitativeGCode} \textbf{Qualitative comparison on GarmentCodeData~\cite{GarmentCodeData:2024}}. From left to right: input (images for ChatGarment~\cite{bian2024chatgarment} and 3D meshes for \method{}), ground truth patterns, and the respective predictions from ChatGarment and \method{} with test-time optimization.}
\end{figure*}
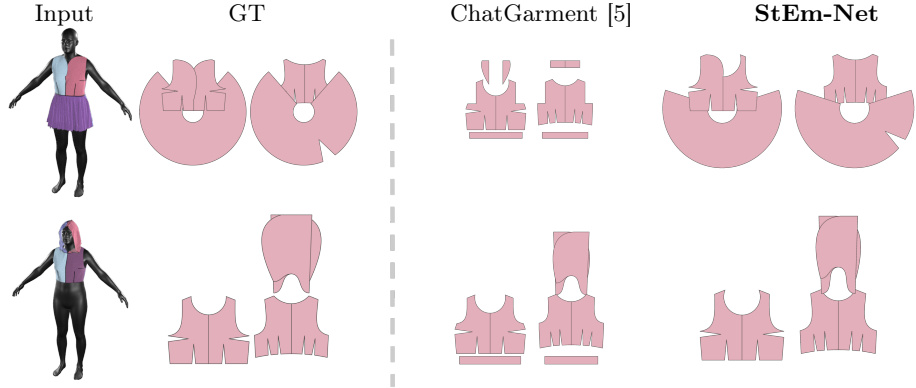

To evaluate the robustness of \method{}, we compare our performance against \textbf{ChatGarment}~\cite{bian2024chatgarment}, a state-of-the-art method for regressing sewing patterns from 2D images, and we report the results in \cref{tab:garmentcode_results}. While \method{} operates directly on 3D point clouds, ChatGarment utilizes a multi-modal approach. To ensure a fair comparison, we provide ChatGarment with 2D images where garment panels are explicitly segmented into distinct patches (as shown in Fig. \ref{fig:qualitativeGCode}).
This comparison highlights a fundamental difference in input representation. ChatGarment relies on visual textures and 2D spatial arrangements to infer the pattern. In contrast, \method{} leverages 3D geometry and surface normals, which provide more definitive cues regarding physical draping and panel connectivity. 

\begin{table}[t]
\centering
\scriptsize
\setlength{\tabcolsep}{4pt}

\begin{adjustbox}{width=0.7\linewidth}
\begin{tabular}{lccccc}
\toprule
\multicolumn{6}{c}{GarmentCodeData TestSet} \\
\midrule
Method 
& Panel L2 $\downarrow$ 
& \#Panels 
& \#Edges 
& Stitch Precision
& Stitch Recall \\
\midrule

ChatGarment \cite{bian2024chatgarment} 
& 10.1
& 12.5\% 
& 45.5\% 
& 24.1\% 
& 25.1\% \\

\method{} 
& 4.4 
& 69.5\% 
& \textbf{82.6}\% 
& \textbf{66.6}\% 
& \textbf{66.4}\% \\

\method{} + Optim 
& \textbf{4.3} 
& \textbf{71.1\%} 
& 80.6\% 
& 64.0\% 
& 64.3\% \\

\bottomrule
\end{tabular}
\end{adjustbox}

\caption{Results on the GarmentCodeData TestSet. The optimization step slightly improves Panel L2 and panel detection, while stitch metrics remain comparable.}
\label{tab:garmentcode_results}
\end{table}

Even though the baseline benefits from pre-segmented patches, which significantly simplify topological inference, our experiments demonstrate that \method{} achieves superior accuracy in parameter regression. We attribute this to the StEm latent space, which preserves the structural relationship between 3D volume and 2D topology more effectively than a 2D-to-2D mapping. Furthermore, our test-time optimization allows \method{} to iteratively refine the pattern based on the input surface, a dynamic capability that purely feed-forward image-based methods lack.

\subsection{Editing from patterns}
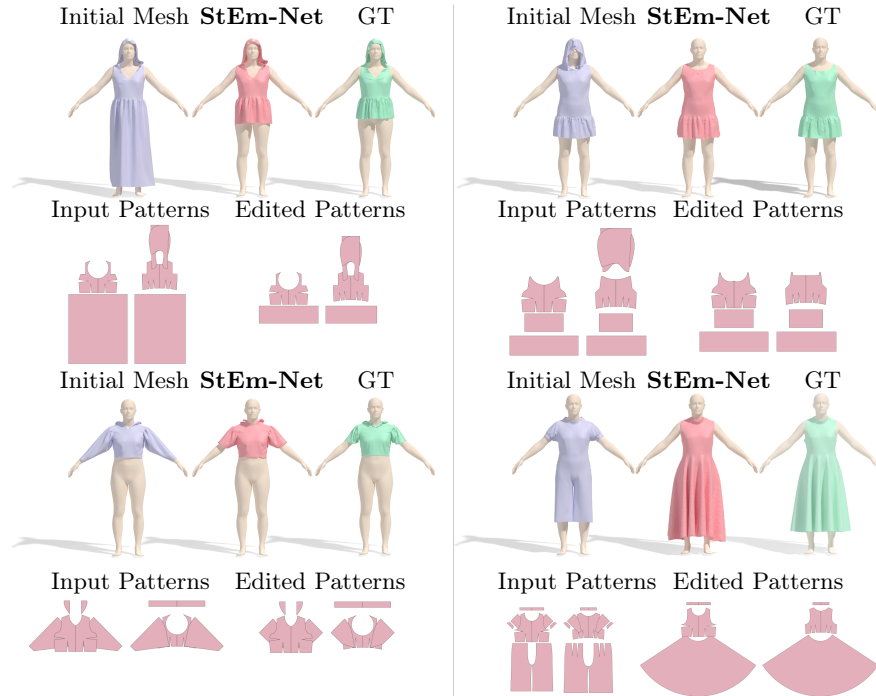
\begin{figure}[t!]
    \centering
       \begin{overpic}[width=0.95\linewidth]{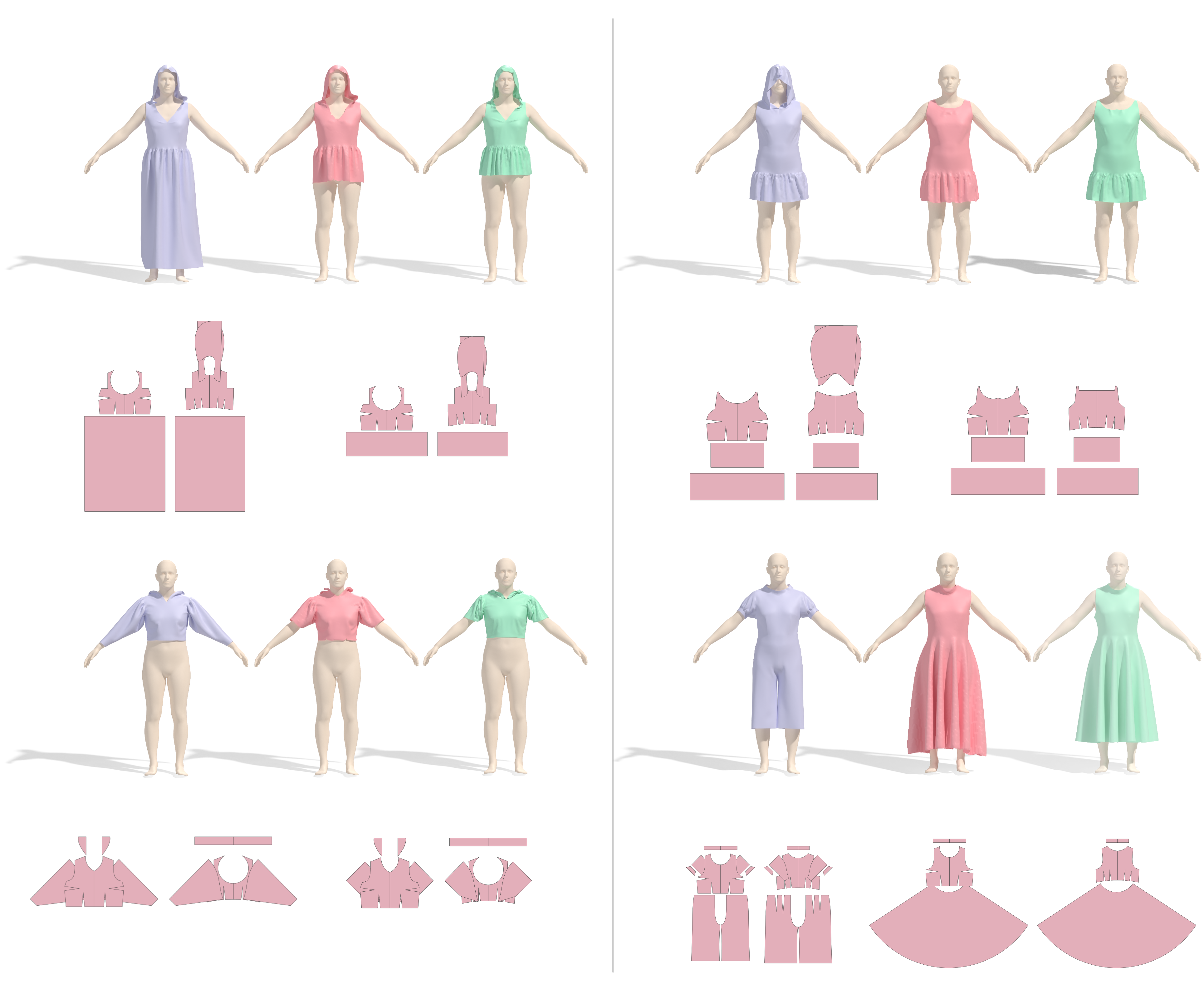}
       
    \put(5,56){Input Patterns}
    \put(26,56){Edited Patterns}
    \put(6,78){Initial Mesh}
    \put(22,78){\textbf{\method{}}}
    \put(40,78){GT}
    
    \put(56,56){Input Patterns}
    \put(76,56){Edited Patterns}
    \put(57,78){Initial Mesh}
    \put(73,78){\textbf{\method{}}}
    \put(91,78){GT}

    \put(5,13.5){Input Patterns}
    \put(26,13.5){Edited Patterns}
    \put(6,36.5){Initial Mesh}
    \put(22,36.5){\textbf{\method{}}}
    \put(40,36.5){GT}
    
    \put(56,13.5){Input Patterns}
    \put(76,13.5){Edited Patterns}
    \put(57,36.5){Initial Mesh}
    \put(73,36.5){\textbf{\method{}}}
    \put(91,36.5){GT}

       \end{overpic}
   \vspace{-0.3cm}
    \caption{Editing examples. For each example, we show on the left the initial sewing pattern with its 3D physical simulation (blue garment). On the right, the edited sewing pattern, with our network's prediction (red) and the ground truth (green). We observe that our framework provides an accurate estimation of the garment's appearance without requiring physical simulation, both for global behaviors (\eg, length of the garment's components) and for the addition or removal of garment parts (\eg, replacing trousers with a skirt or removing sleeves and the hood).}
    \label{fig:editing}
\end{figure}

A key advantage of our shared latent space is the ability to perform 3D garment editing via 2D pattern manipulation, without requiring a physical simulator. By initializing a garment in a specific configuration, users can explicitly modify the 2D sewing pattern parameters and observe the corresponding 3D drape through the forward pass of our neural simulator ($\mathcal{E}_{bm} \to \mathcal{D}_{3D}$). \method{} enables a seamless mapping from parametric 2D edits to high-fidelity 3D mesh updates. This provides a fast, physics-free alternative to traditional CAD pipelines, enabling intuitive, interactive garment design. During the edit, we assume the input normals do not change. Even if such an approximation produces sub-optimal input, our results show that the network is robust to it and operates normally. We report examples of our editing in \cref{fig:editing}.

\subsection{Generalization to Real Scans}

Although trained exclusively on curated synthetic garments, we evaluate our method's ability to generalize to real-world scans—a highly challenging setting due to inherent sensor noise and topological inaccuracies. To process this data, we utilize scans from the 4DDress and CloSe datasets, specifically selecting instances with wide arm and leg stances. We first unpose the scans to a canonical A-pose and leverage the dataset's segmentation labels to isolate individual garments before inference.

\begin{figure}[t!]
  \centering
  % ---- Left: mesh comparison ----
  \begin{minipage}[t]{0.48\linewidth}
    \centering
    {\raggedleft
    \begin{overpic}[width=0.85\linewidth, percent]{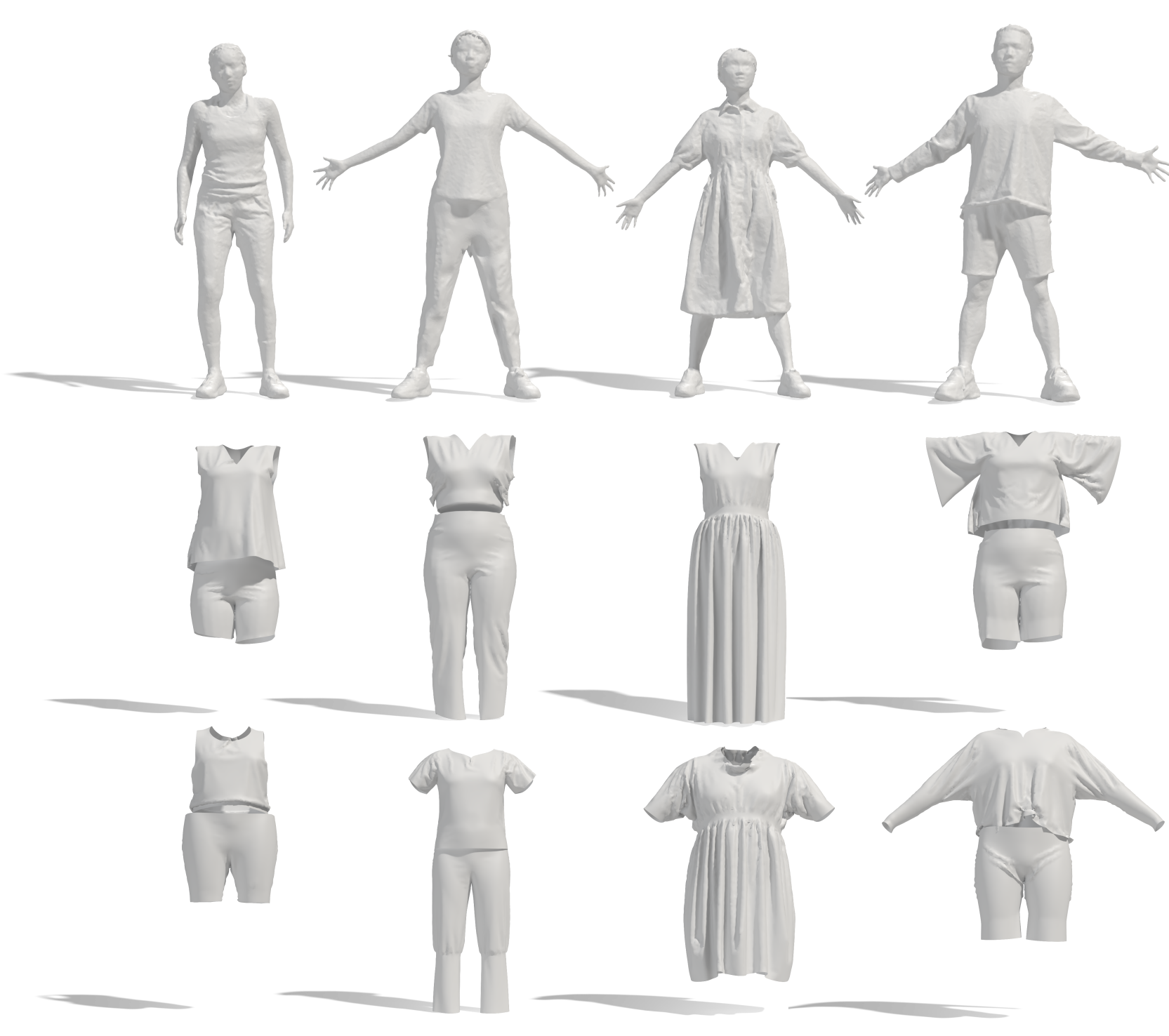}
    \scriptsize
        \put(-16,70){\selectfont Input}
        \put(-16,65){\selectfont Mesh}
        \put(-16,39){\selectfont Neural}
        \put(-15,34){\selectfont Tailor}
        \put(-20,15){\selectfont StEm-Net}
    \end{overpic}\par}
    \vspace{-0.1cm}
    \captionof{figure}{Qualitative comparison on scans from 4DDress. StEm-Net reconstructs garments that adhere more closely to the input geometry than NeuralTailor.}
    \label{fig:scans_results}
  \end{minipage}
  \hfill
  % ---- Right: noise study ----
  \begin{minipage}[t]{0.48\linewidth}
    \centering
    {\raggedleft
    \raisebox{0.4cm}{%   <-- positive raises up; increase to lift more
    \begin{overpic}[width=0.90\linewidth, percent]{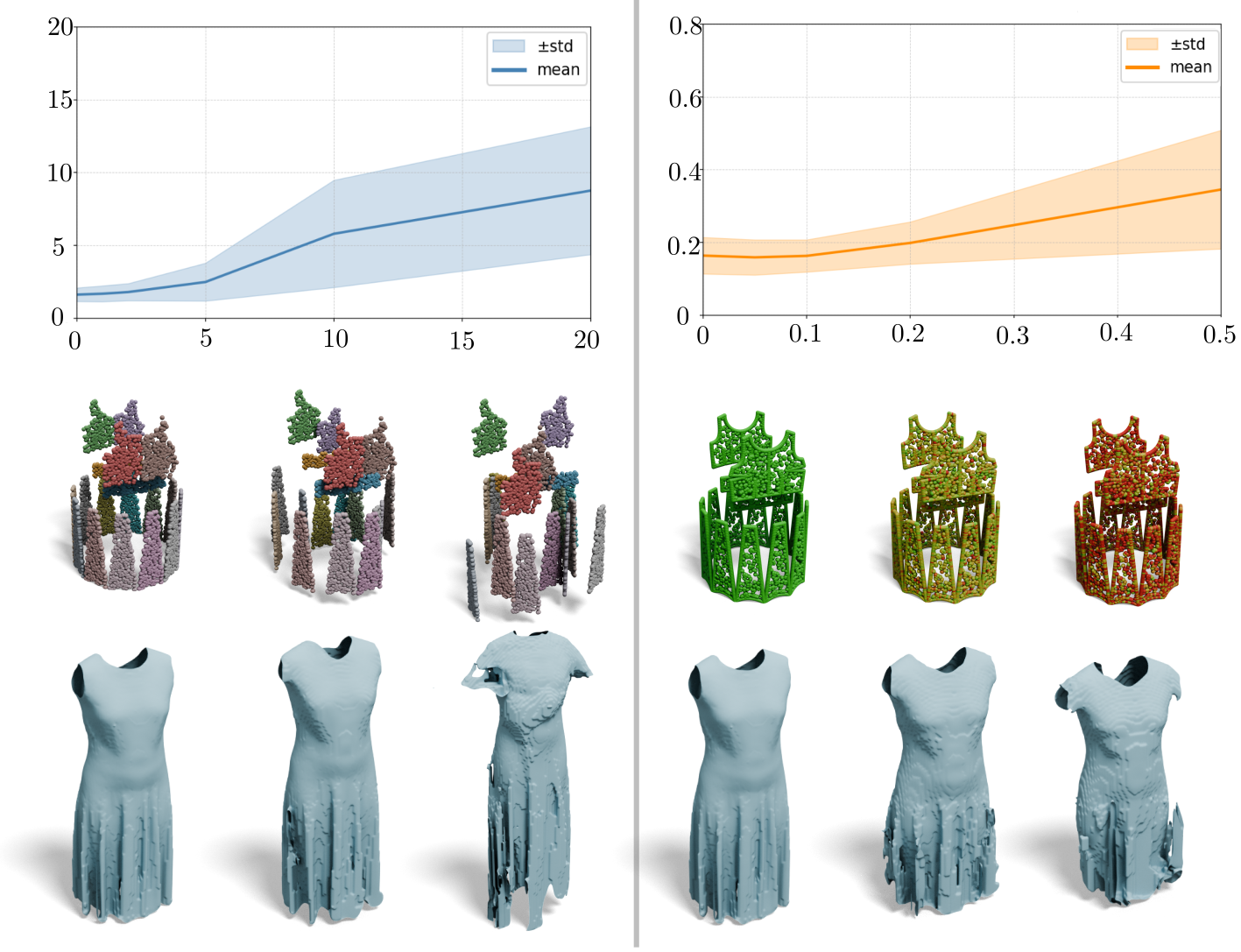}
    \scriptsize
        \put(14,47){\scalebox{0.6}{\tiny Translation Noise $\sigma$ (cm)}}
        \put(70,47){\scalebox{0.6}{\tiny Normal Noise $\sigma$}}
        \put(2,50){\scalebox{0.6}{\rotatebox{90}{\tiny Chamfer Distance (cm)}}}
        \put(52,50){\scalebox{0.6}{\rotatebox{90}{\tiny Chamfer Distance (cm)}}}
        
        \put(-9,34){Input}
        \put(-7,13){$\mathcal{D}_{3D}$}
        \put(8,-3){$\sigma_0$}
        \put(23,-3){$\sigma_{5}$}
        \put(40,-3){$\sigma_{10}$}
        
        \put(56,-3){$\sigma_{0.0}$}
        \put(72,-3){$\sigma_{0.25}$}
        \put(88,-3){$\sigma_{0.50}$}
    \end{overpic}%
    }\par}
    \vspace{-0.1cm}
    \captionof{figure}{We report the chamfer distance degradation with different Gaussian noise levels for BoxMesh translation (left) and normal orientation (right). Point color indicates the panel on the left, and the amount of per-point noise on the right.}
    \label{fig:noise}
  \end{minipage}
\end{figure}

As shown in \cref{fig:scans_results}, our method accurately reconstructs 3D garments from raw scan data, demonstrating its practical viability for real-world digital fabrication and streamlining garment acquisition pipelines. Furthermore, we provide a quantitative comparison against NeuralTailor \cite{NeuralTailor2022} in Table \ref{tab:scans}, measuring the Chamfer Distance between the predicted pattern and the scan on the CloSe samples. Our approach significantly outperforms the baseline across all metrics.

\begin{table}[t!]
    \centering
    \small
    \begin{tabular}{lccc}
    \toprule
     & CD$_{STP}$ & CD$_{PTS}$ & CD$_{bidir}$ \\
    \midrule
    NeuralTailor & 2.54 & 1.26 & 1.90 \\
    StEm-Net + Optim & \textbf{0.96} & \textbf{0.67} & \textbf{0.81} \\
    
    \bottomrule
    \end{tabular}
    \caption{Mean Chamfer Distance for \textbf{S}can-\textbf{T}o-\textbf{P}attern, \textbf{P}attern-\textbf{T}o-\textbf{S}can, and Bidirectional Chamfer Distance (CD). We collected 543 CloSe scans and reposed them to the canonical pose. 
    }
    \label{tab:scans}
\end{table}

\subsection{Ablation on Input Normals}

To evaluate our model's robustness, we analyze the 3D reconstruction error under varying levels of Gaussian noise applied to the input BoxMesh. As illustrated in \cref{fig:noise}, we introduce noise to both the panel displacements (left) and the normal directions (right). In both scenarios, our method demonstrates significant resilience; the reconstruction error increases sublinearly relative to the noise level, indicating that the model remains stable and does not strictly overfit to pristine synthetic inputs.

%% file: sections/07_Conclusions.tex
In this work, we introduce \method{}, a novel framework that, for the first time, unifies 2D sewing patterns and 3D garments in a common latent space, here called Stitched Embeddings. This is achieved by representing sewing patterns as BoxMeshes, thereby making their connection with the 3D garment more direct and simpler to learn. Our method is fully differentiable, end-to-end trained, and does not require a physical simulator for test-time optimization. We showcase our advantages on two different datasets, improving the state of the art in sewing pattern reconstruction from 3D point clouds. Our framework supports test-time adaptation via backpropagation, making it highly computationally efficient. Within the same framework, we also enable direct propagation of sewing pattern edits to the 3D mesh, providing a practical, flexible tool for manufacturing. While our results are still not fully aligned with those from physical simulation, we view our work as an important milestone toward scalable digital garments, closing the gap between reconstruction and tailoring pipelines. Scaling to multi-layered attire, different materials, varying body models, and fine-grained nuances, such as pockets or ethnic garments, would be an interesting follow-up to our work. The principal obstacle in this direction is the scarcity of high-quality garment datasets.

\textbf{\emph{Acknowledgements.}} 
We thank the whole RVH team for the support. The authors thank the International Max Planck Research School for Intelligent Systems (IMPRS-IS) for supporting Andrea Sanchietti. This project was supported by Meta. Gerard Pons-Moll was supported by the German Federal Ministry of Education and Research (BMBF): Tübingen AI Center, FKZ:01IS18039A, by the Deutsche Forschungsgemeinschaft (DFG, German Research Foundation) - 409792180 (Emmy Noether Programme, project: Real Virtual Humans). GPM is a member of the Machine Learning Cluster of Excellence, EXC number 2064/1 - Project number 390727645, and is supported by the Carl Zeiss Foundation.